\UseRawInputEncoding
\pdfoutput=1

\documentclass[journal]{IEEEtran}
\usepackage{graphicx}

\usepackage{tikz}
\usepackage{comment}
\usepackage{amsmath,amsfonts} 
\usepackage{color}
\usepackage{cite}
\usepackage{subfigure}
\usepackage {multirow}
\usepackage[caption=false,font=normalsize,labelfont=sf,textfont=sf]{subfig}

\begin{document}

\newcommand{\etal}{\textit{et al}.}
\newcommand{\etc}{\textit{etc}}
\newcommand{\ie}{\textit{i}.\textit{e}.}
\newcommand{\eg}{\textit{e}.\textit{g}.}

\title{DialogueNeRF: Towards Realistic Avatar Face-to-Face Conversation Video Generation}



%
\author{Yichao Yan, Zanwei Zhou, Zi Wang, Jingnan Gao, Xiaokang Yang,~\IEEEmembership{Fellow,~IEEE}
\thanks{Yichao Yan, Zanwei Zhou, Zi Wang, Jingnan Gao, and Xiaokang Yang are with MoE Key Lab of Artificial Intelligence, AI Institute, Shanghai Jiao Tong University, Shanghai, China. E-mail: \{yanyichao, SJTU19zzw, w4ngz1,  gjn0310, xkyang\}sjtu.edu.cn. 

Yichao Yan and Zanwei Zhou contributed equally to this paper.} 
}
%
%
\maketitle

\begin{abstract}
Conversation is an essential component of virtual avatar activities in the metaverse. With the development of natural language processing, textual and vocal conversation generation has achieved a significant breakthrough. 
However, face-to-face conversations account for the vast majority of daily conversations, while most existing methods focused on single-person talking head generation.
In this work, we take a step further and consider generating realistic face-to-face conversation videos. Conversation generation is more challenging than single-person talking head generation, since it not only requires generating photo-realistic individual talking heads but also demands the listener to respond to the speaker.
In this paper, we propose a novel unified framework based on neural radiance field (NeRF) to address this task.
Specifically, we model both the speaker and listener with a NeRF framework, with different conditions to control individual expressions. The speaker is driven by the audio signal, while the response of the listener depends on both visual and acoustic information. In this way, face-to-face conversation videos are generated between human avatars, with all the interlocutors modeled within the same network.
Moreover, to facilitate future research on this task, we collect a new human conversation dataset containing 34 clips of videos.
Quantitative and qualitative experiments evaluate our method in different aspects, \eg, image quality, pose sequence trend, and naturalness of the rendering videos. Experimental results demonstrate that the avatars in the resulting videos are able to perform a realistic conversation, and maintain individual styles. All the code, data, and models will be made publicly available.
\end{abstract}

\section{Introduction}

The term metaverse was first proposed by Neal Stephenson in his science fiction novel \textit{Snow Crash} published in 1992, where humans can interact with each other as avatars. 
Nowadays, with the development of computer graphics and computer vision, the metaverse is no longer a dream in novels but an accessible thing in our daily lives. 
One of the most exciting applications in the metaverse is that humans can have their virtual avatars to communicate with others via chats, meetings, conferences, \etc. These engaging application scenarios motivate us to explore the conversations between humans, and then try to build a realistic human avatar conversation process in the metaverse.

Conversation is an essential part of human social activities. 
The way people communicate can be mainly divided into three categories: textual communication, vocal communication, and face-to-face communication. In the area of natural language processing (NLP), there have been some classic methods that are able to generate realistic textual communication processes~\cite{kumar2016ask, luo-etal-2018-auto, wang-etal-2018-learning-ask}, and combine speech synthesis models~\cite{oord2016wavenet, wang17n_interspeech} to generate realistic vocal communication processes. Some textual conversation datasets
~\cite{lison-tiedemann-2016-opensubtitles2016, lowe-etal-2015-ubuntu, danescu-niculescu-mizil-lee-2011-chameleons} have been constructed as well. While the textual and vocal conversation process generation has achieved significant success, human face-to-face conversation generation has not been seriously considered. 
Human communication is a complex joint activity~\cite{r39}. Some early works~\cite{r41} focusing on face-to-face human interaction find that conversational partners tend to receive some subtle expression information from the other side and exhibit increasingly similar behavior throughout a conversation.
Therefore, the ability of virtual avatars to naturally interact with each other is of great significance for a realistic conversation process generation in metaverse, and their broad application scenarios inspire us to rediscover this rarely explored field from the perspective of image processing. 

\begin{figure}[t]
  \centering
   \includegraphics[width=\linewidth]{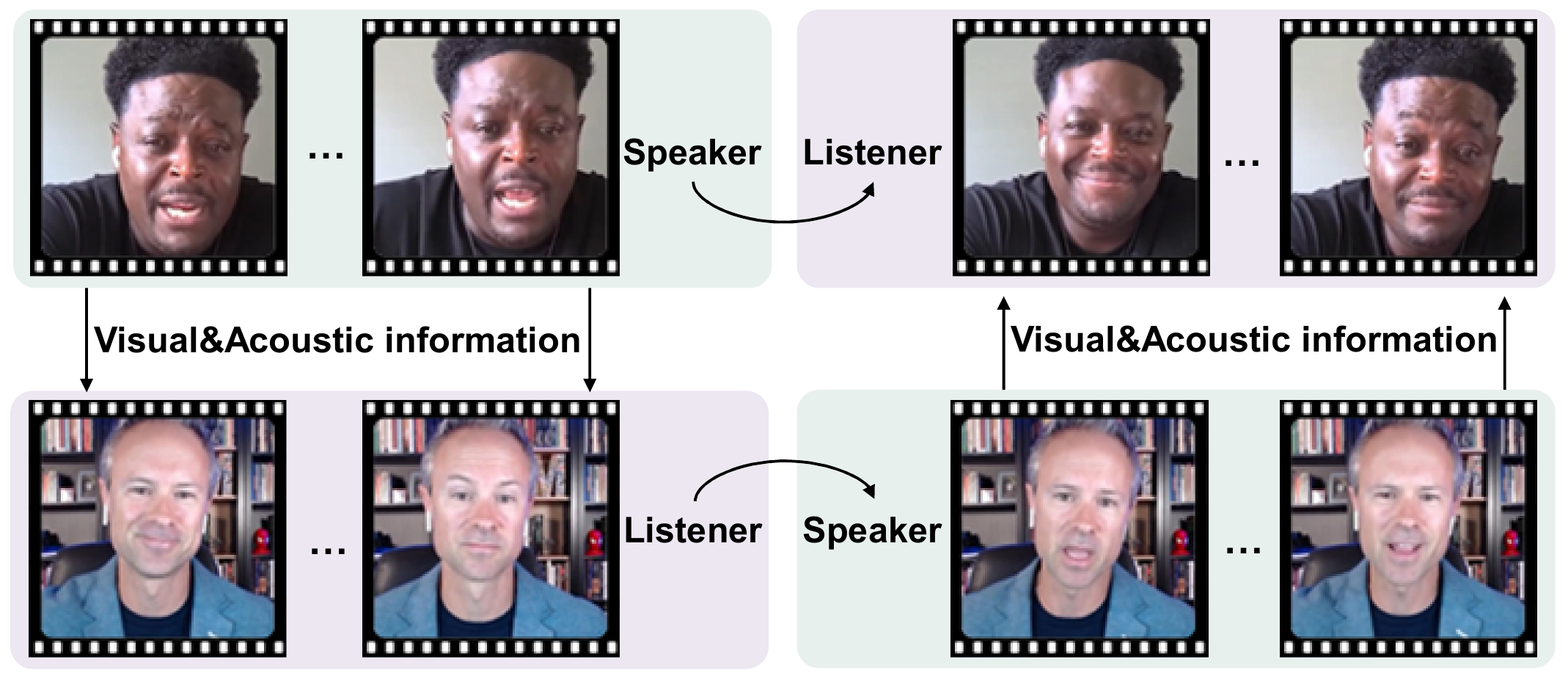}

   \caption{Schematic diagram of our proposed task. In a conversation process, typically, there are two types of roles, \ie, the speaker and listener. The speaker transmits both visual and acoustic information to the listener. Consequently, the listener will respond to the information and take some responsive actions like nodding and smiling. The role of the speaker and listener can switch during the conversation.}
   \label{fig:conversation process}
\end{figure}

Different from textual and vocal conversations that only have a single source of information (text and voice), interlocutors in face-to-face conversations receive and transfer diverse visual and acoustic information to each other.
Therefore, to model the conversation process, it is necessary to exploit the multi-modal signals from all the interlocutors, \eg, audio, head pose, and expression. 
Moreover, to render the avatars in face-to-face conversations, we focus on a series of methods~\cite{r3, r1, r4, r5, r6, r7, r8, r9} that are able to generate dynamic avatars. It is straightforward to reconstruct and manipulate a parametric 3D face model~\cite{r1,r3,r4,DBLP:journals/tip/GuoCZ21,DBLP:journals/tip/JiangZDLL18,DBLP:journals/tip/ChenWWSLB20,DBLP:journals/tip/YangCCCW23,DBLP:conf/iccv/RichardZWTS21} to build a facial avatar, but the avatars synthesized by these methods cannot be as realistic as the real humans. Some generative adversarial network (GAN) based techniques~\cite{r21, zhang2021flow, r7} build talking avatars via audio-to-face translation. But they are hungry for numerous and diverse data and high computing resources.
Some recent works~\cite{r8, r9} use neural radiance fields (NeRF)~\cite{r12} to reconstruct a dynamic facial avatar. For example, Guo \etal~\cite{r8} presented an audio-driven neural radiance fields model (AD-NeRF). 
However, it only constructs a single person using a NeRF model, and thus is difficult to scale to multiple avatars with limited memory overheads. Additionally, AD-NeRF can only imitate the head poses from existing videos and is unable to generate new sequences of head poses maintaining the corresponding individual's characteristics. In a nutshell, all these methods above focus on building one independent speaking avatar and are unsuitable for modeling a responsive listening avatar.

In this paper, we aim to generate a natural and photo-realistic face-to-face conversation process between virtual avatars, as shown in Fig.~\ref{fig:conversation process}. 
Our goal is
to generate two specific human conversation video with only a sequence of speaking audio.
However, it is hard to obtain an available dataset suitable for our task since most of the existing datasets only contain the speaking persons, while the whole conversation processes (with responsive listeners) are not included.
To address this issue, we collect a corresponding dataset collected from public websites. The dataset contains 34 video clips, which have a total duration of 2200s. There are two individuals in one clip, one is speaking and the other is listening.
It contains the listener's natural responses to the speaker, including nodding, eye blinking, smiling, \etc. 
To the best of our knowledge, we are the first to address this task with a novel dataset and a primary method.

To tackle this task, we propose an approach termed DialogueNeRF, which is a unified framework for conversation video generation. Specifically, we simultaneously model the speaker and the listener with a NeRF-based~\cite{r12} framework, where the interlocutors are controlled by two latent codes, respectively. Given an audio signal, we extract the disentangled audio-face features to generate a lip-synchronized speaker. In order to reflect the influence of the speaker on the listener, we supply the speaker's audio feature to the listener as conditions. Meanwhile, a time-series-forecasting model is designed to generate head pose sequences for both the speaker and listener, where the generated pose sequences correspond to the speaker's audio. These multi-modal signals will be utilized to synthesize the corresponding avatars via a code-controlled neural radiance fields network.
Our model then simultaneously builds various avatars that possess diverse heads and torsos, by varying the control latent codes and multi-modal conversation signals.
Since the conversation scenarios may contain violent shakes in the torsos, we build a deformation field, such that the movements can be more smooth. 


\textbf{In summary, our contributions are:}
\begin{itemize}
\setlength{\itemsep}{0pt}
 \setlength{\leftmargin}{-15pt}
    \item We present a brand new task that focuses on synthesizing a realistic human face-to-face conversation process. It is essential for realistic communication between virtual avatars in future metaverse applications. 

    \item We collected and (will) release a new dataset containing 34 conversation video clips for realistic human face-to-face conversation generation. To our best knowledge, this is the first dataset focusing on a face-to-face human communication process. In contrast, existing public datasets restrictively contain a single person.

    \item We present a dedicatedly designed framework named DialogueNeRF, to generate human avatars for face-to-face interaction. To our best knowledge, our method first achieves modeling and rendering the interlocutors in one model.
    With minutes of conversation videos needed for training, we can obtain realistic conversation processes through rendering.
    The source code will be made publicly available.
\end{itemize}

\section{Related Work}

\textbf{Speaker Diarization} is a technical process that splits up an audio recording stream that often includes several speakers into homogeneous segments, \ie, a task to find ``who speak when''. The segments correspond to individual speakers.
Chopping up an audio recording file into shorter, single-speaker \cite{anguera2012speaker} segments includes several following steps:
\textbf{1)} Voice activity detection: separating speech from background noise in the audio recording or stream and discarding non-speech parts;
\textbf{2)} Speech segmentation: producing small segments of an audio file by conducting speaker switch points detection and overlapped speech detection, \etc;
\textbf{3)} Embedding extraction: putting all the embedded speech segments created and collected in the previous step, and then creating a neural network for those segments;
\textbf{4)} Clustering embeddings: grouping the embeddings according to the actual identity;
\textbf{5)} Labeling clusters: 
once the clusters are appropriately labeled, the audio can then be segmented into individual clips for each speaker.
Open-source tool-kits such as S4D~\cite{r14}, Kaldi~\cite{r15}, ALIZE~\cite{r16}, pyAudioAnalysis~\cite{r17}, and Pyannote~\cite{r18} all provide powerful functions for speaker diarization. Pyannote typically provides an end-to-end neural implementation for the speaker diarization, as well as plenty of pre-trained speaker diarization models.

\begin{table*}[tb!]
\small
\centering
\caption{Statistics of our collected dataset.}
\begin{tabular}{c|c|c|c|c|c|c}
\hline
~Total Duration~ & ~\#Clips~ & \#Nods & ~Avg. Duration~ & Avg. Nods & Resolution & ~\#Individuals~ \\ \hline
2200s        & 34      & 1409(times)      & 64.7s       & 38(times/min)    & 550 $\times$ 550 and  385 $\times$ 385    & 11            \\ \hline
\end{tabular}
\label{tab:data_stat}
\end{table*}

\begin{figure*}[t]
  \centering
   \includegraphics[width=\linewidth]{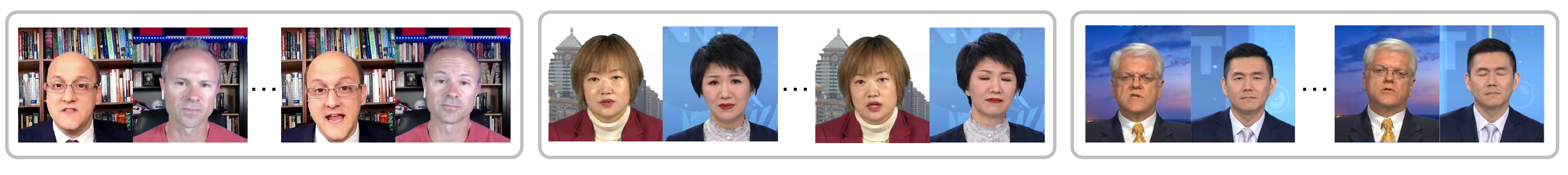}
   \caption{Examples of our dataset collected from a public website. The left individuals are speakers and the right  are listeners. The face regions of speaker-listener pairs are cropped from the original videos manually.}
   \label{fig:data}
\end{figure*}

\textbf{Implicit Neural Scene Networks} utilizes neural networks as implicit functions to represent the shape and appearance of the scenes. Scene representation networks (SRN) \cite{r27} is proposed to represent a continuous, 3D-structure-aware scene by encoding both geometry and appearance. Neural Radiance Fields (NeRF) \cite{r12} uses an underlying implicit representation of the object to render novel view synthesis of complex scenes and achieves state-of-the-art rendering quality. 
Then many NeRF-based methods are proposed for different tasks.
Schwarz \etal~\cite{r19} integrate GAN method into NeRF model and propose a multi-scale patch-based discriminator for radiance fields. Niemeyer \etal~\cite{r28} disentangle the objects from the background as well as individual object shapes and appearances by representing scenes as the compositional generative model, thus improve the editability of the synthesized scenes. There are other concurrent works that learn deformable objects with additional latent codes. For example, Peng \etal~\cite{peng2021neural} add human pose parameters as additional input to synthesize deformable human body. Park \etal~\cite{r10} fits the deformable scene into a hyper-space to address the discontinuity in the deformation.

\textbf{Audio-driven Talking Head Generation} aims to synthesize a photo-realistic talking head video, driven by a piece of audio or motion sequence. 
This task is very useful in many interactive applications including video conferencing, customized services, news broadcasting, \etc. 
Therefore, the research community currently also show great interest in this task, aiming to reenact a specific person in sync with arbitrary input speech sequences. 
Specifically, there are two categories of methods to handle this issue: 2D-based methods and 3D-based methods. 
2D-based method commonly synthesize the video by adopting landmarks, semantic maps, \etc. A number of works try to generate videos of identity-independent videos \cite{r23, r8, r6, r24, r13, r25, r26}. Chen \etal~\cite{r8} and Yu \etal~\cite{r24} utilize facial landmarks as the internal representation to bridge audio and facial images. Zhou \etal~\cite{r6} predict spontaneous head poses alongside the audio. Song \etal~\cite{r25}, Vougioukas \etal~\cite{r26} and Prajwal \etal~\cite{r13} significantly improve lip-syncing performance by adopting GAN methods.
3D-based methods are better at motion controlling, but the drawback is the high cost in constructing 3D models. Some early methods like Suwajanakorn \etal~\cite{r21} and Karras \etal~\cite{r22} adopt a pre-built 3D face model of a specific person, and then drive the model by learning a sequence (a piece of audio) to sequence (motions of 3D model) mapping. Recent methods tend to reconstruct the 3D face model out of training data. Thies \etal~\cite{r7} employ a latent 3D face space in the deep neural network to reenact a specific person. 
AD-NeRF \cite{r8} trains two conditional neural radiance fields (NeRFs) to render the head and torso parts respectively. AD-NeRF also integrates the audio features into the implicit neural function.

\begin{figure}[t]
  \centering
   \includegraphics[width=\linewidth]{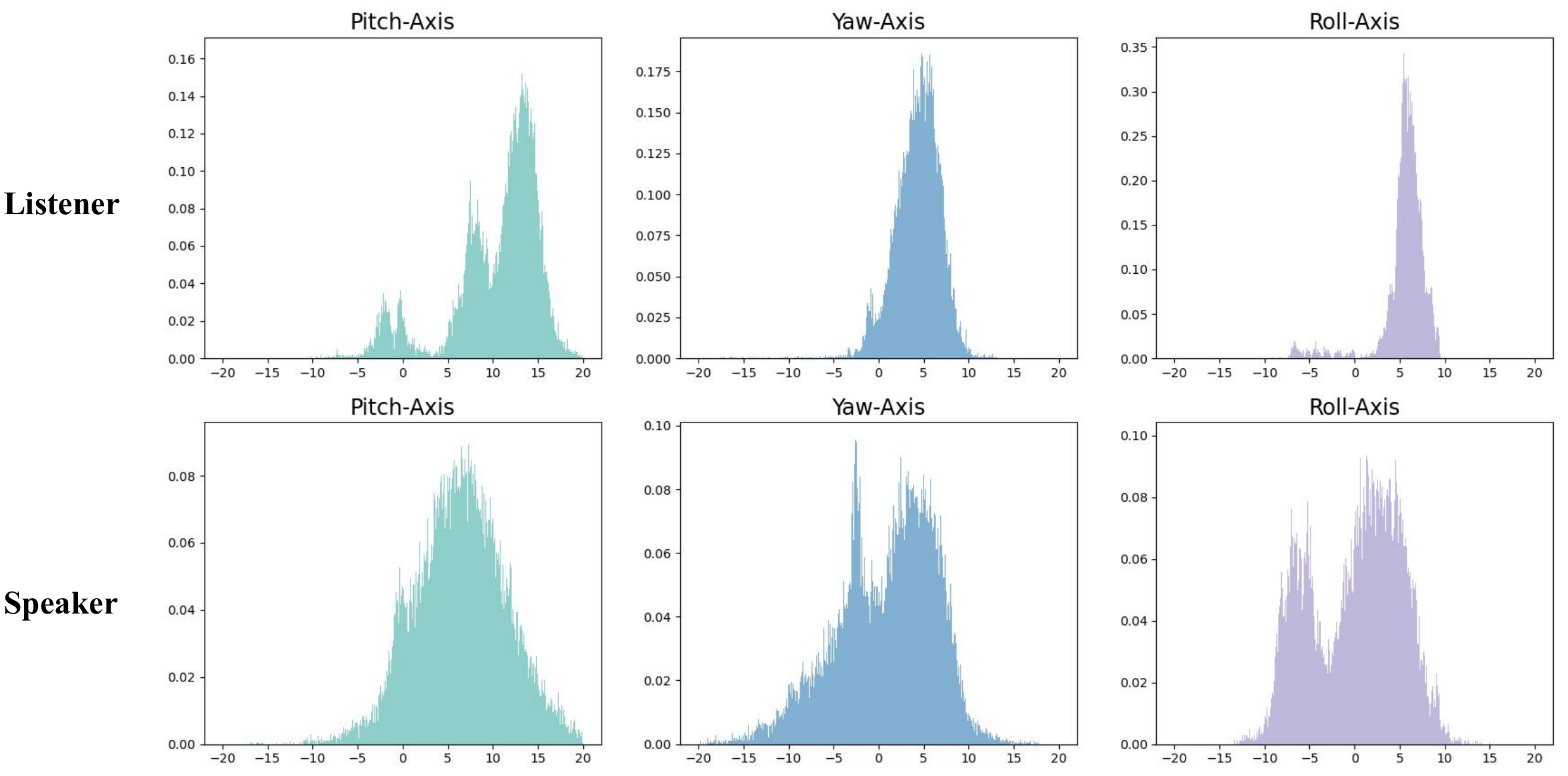}
\vspace{-5pt}
   \caption{The statistics on head pose Euler angles of listeners and speakers, respectively. }
   \label{fig:euler_hist}
\end{figure}

\begin{figure}[t]
  \centering
   \includegraphics[width=\linewidth]{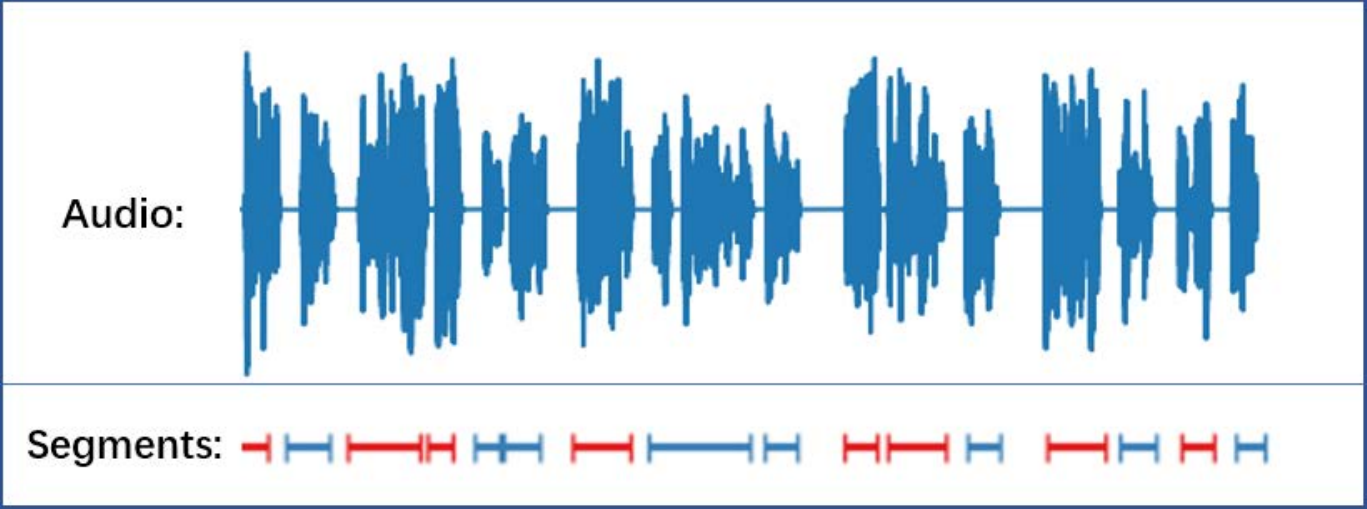}
   \caption{The result of two individuals' speaker diarization. The audio has been separated into many segments. Blue lines and red lines in segments represent different speakers.}
   \label{fig:speaker diarization}
\end{figure}

\section{Face-to-face Virtual Conversation}


\begin{figure*}[t]
    \centering
  \includegraphics[width=1.\linewidth]{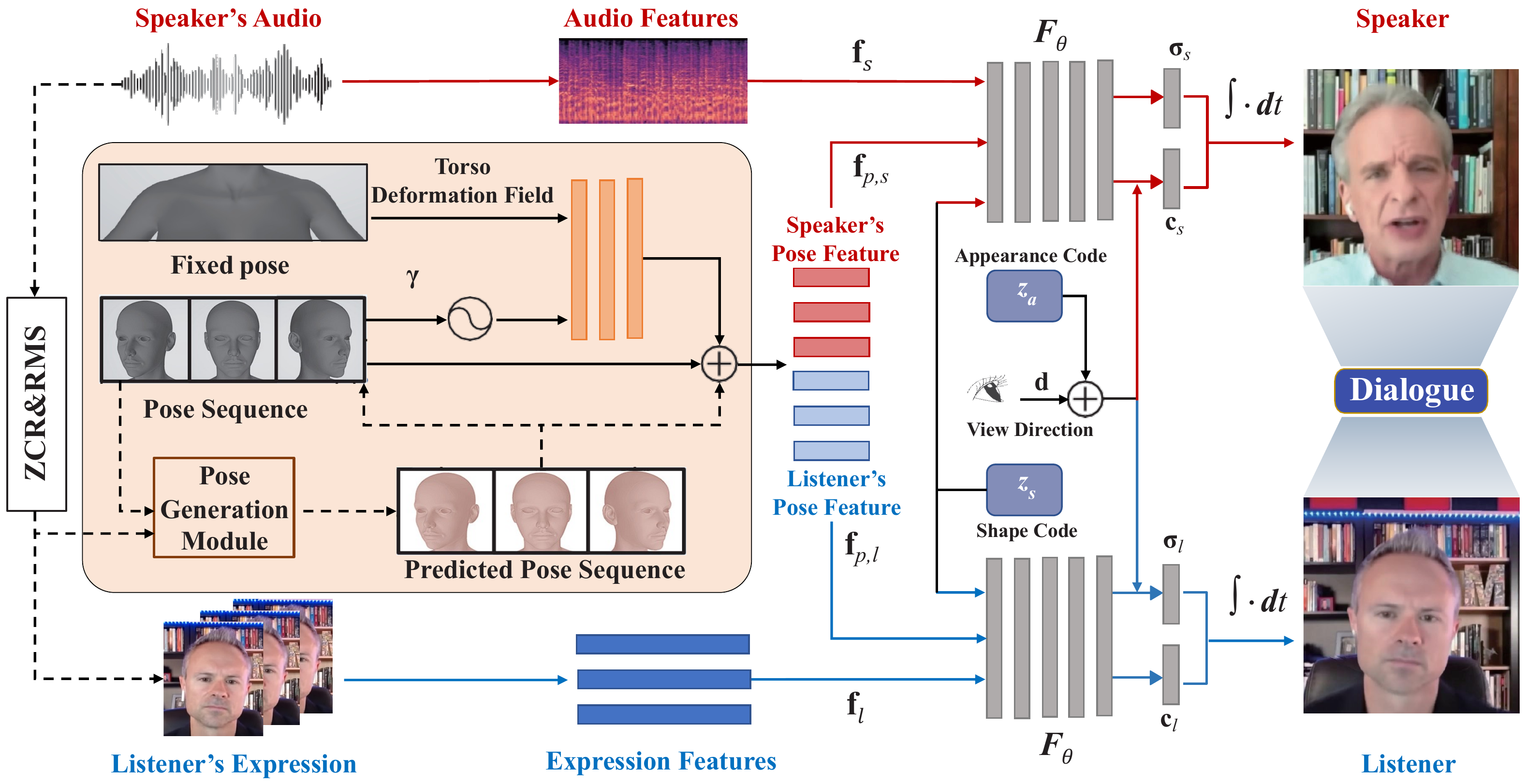}
\vspace{-10pt}
  \caption{The framework of our DialogueNeRF. First, the pose sequence is extracted from the video dataset and used to supervise the training of the pose generation module and rendering module. Then, the audio features and expression features are extracted and disentangled
  respectively for the speaker and listener ($\mathbf{f}_{s}$ and $\mathbf{f}_{l}$). During inference, we use the pose generation module (specifically TCN~\cite{bai2018empirical}) to generate predicted pose sequences ($\mathbf{f}_{p,s}$ and $\mathbf{f}_{p,l}$). These above signals will then be used to synthesize the corresponding avatars via neural radiance fields. 
  The rendering module is a conditional dynamic neural radiance field. It is controlled by two latent codes ($\mathbf{z}_{s}$ and $\mathbf{z}_{a}$) determining the synthesis of shape and appearance, respectively. With different latent codes, our model is capable of building various avatars that possess diverse heads and torsos within the same framework.
  }
  \label{fig:framework}
\end{figure*}

\subsection{Problem Formulation }
In this paper, we focus on human face-to-face conversation generation.
In most cases, conversation participants can be divided into two types of roles: \emph{speaker} and \emph{listener}. Usually, one participant can only play one role at a particular time, and the same participant can play different roles at different periods. A face-to-face conversation involves an interchange of visual and acoustic messages between the speaker and listener.

The objective of our task is to generate human face-to-face virtual conversations with limited inputs, \ie, simultaneously generating a video containing the talking speaker and responding listener given an audio sequence. 
The generation of virtual conversations is a cross-modal problem, which takes both visual and acoustic information exchange into account. Some works~\cite{r41} discover that conversational partners tend to receive subtle information from the other side. However, since the listener's feedback is weak and difficult to simulate, for the sake of simplicity, in virtual conversations, we mainly consider the influence from speaker to listener. Therefore, the target virtual conversation that we expect to generate can be described as a bilateral or multilateral face-to-face interactive process, where each avatar participant transmits visual and acoustic information to others at a certain period as a speaker, and switches to receive the cross-modal information from the speaker at the next period as a listener, as shown in Fig.~\ref{fig:conversation process}.


\subsection{Our Collected Dataset}
Face-to-face conversations are pretty common in our daily lives, especially in video conferences, online interviews, and so on. 
However, there are no suitable datasets for this area. 
Some previous works focusing on conversations only provide textual or vocal conversation data.
Some datasets containing talking persons are composed of single individuals. These kinds of datasets lack conversational procedures. 
Thus it is indispensable to explore this area and propose a dataset for the community to solve this issue jointly.

We collect a dataset containing 34 video clips from YouTube and CGTN, which have some real conversations and interviews, including both formal and informal situations. The collected clips should satisfy several conditions: 1) Two individuals appear in one clip simultaneously. This ensures that the listener's responses are consistent with the information provided by the speaker. 2) The face and body are not covered. It is beneficial to the quality of generated talking heads. Following these principles, we edited the original videos to fetch some clips. The image resolution is 550 $\times$ 550 for the YouTube videos and 385 $\times$ 385 for the CGTN videos.
In addition, the duration of one clip ranges from 24 seconds to 182 seconds, which is sufficient for training a talking head generation model. Fig. \ref{fig:data} illustrates some examples of our collected dataset. Table \ref{tab:data_stat} shows the statistics of our dataset.

\begin{figure*}[t]
    \centering
  \includegraphics[width=1.\linewidth]{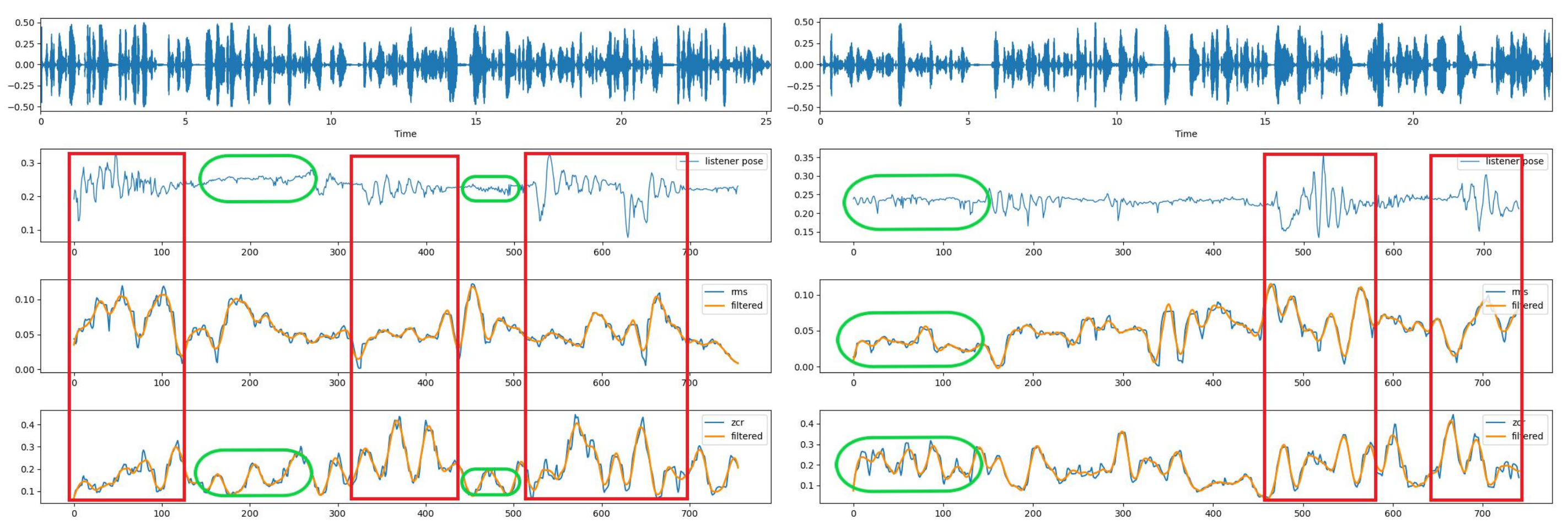}
  \caption{The weak correlation between ZCR, RMS, and the head pose sequences of listeners. We use   Euler angle values on the pitch axis to represent the listeners' responses (particularly nodding) to speakers. Intuitively, it can be found that usually, the sections of ZCR and RMS with gentle undulations and small values correspond to the relatively smooth sections of poses (shown in green capsules), and sections of ZCR and RMS with drastic changes and large values correspond to the fluctuating sections of poses (shown in red boxes). }
  \label{fig:weak_correlation}
\end{figure*}


In Fig. \ref{fig:euler_hist}, we visualize the distribution of the head pose Euler angles of the speakers and listeners, respectively. It can be observed that during the conversation process, the speakers generally have larger head pose variations than the listeners. This may be due to that the listeners generally have limited repetitive responsive patterns, \eg, nodding, smiling, shaking heads, \etc.
For the videos that contain two people talking alternately, we use a speaker diarization tool~\cite{r18} to separate the video into two parts, as shown in Fig. \ref{fig:speaker diarization}. After manually checking the diarization results, then we are able to distinguish the role (speaker or listener) of the corresponding individuals in the training process. 
This dataset contains no sensitive information and would be released only for research purposes under restricted licenses.


\section{DialogueNeRF}
Our framework is shown in Fig. \ref{fig:framework}. Since we divide a virtual conversation process into different periods, we can model the conversation fragment by fragment. First, the multi-modal signals like audio features, head poses, and implicit expression features are generated by the corresponding modules. 
Second, according to the state of the avatars, we provide signals of different modals as conditions. Different avatars are synthesized in the same network and controlled by different latent codes. Then visual avatars are rendered from neural radiance fields using volume rendering. In the training stage, we extract the head pose sequences from individuals in the training videos. The head poses are then used as ground truth to supervise the training of the pose generation module and as camera parameters for visual avatar synthesis. In the inference stage, the generated head pose sequences will replace the extracted head poses to be the input of the pose generation module and rendering module.

\subsection{Multi-modal Condition Generation}
\label{subsec:Modal Signal Generation}
Synthesizing a face-to-face conversation requires some signals to control the virtual avatars.
We mainly focus on three types of signals: audio features, head poses, and implicit expression. In this section, we  introduce how to generate different kinds of signals as the conditions for generation in our framework.

\textbf{Audio Feature Extraction.}
In order to obtain the meaningful audio feature consistent with the speaking avatars' mouth shapes, we employ a powerful audio feature encoder from the pre-trained {Wav2Lip} model~\cite{r13} to extract a 512-dimensional feature code from each 200 ms audio clip. According to our observation, this audio feature shows a stronger correlation to the speaker's mouth shape compared to {DeepSpeech}~\cite{r35} used in AD-NeRF~\cite{r8}. As a result, the generated mouth can open wider and show richer variations. As for acoustic information used for pose generation, the utilized audio features are the same for speakers and listeners. We find that in many circumstances, the listener's responses become more intense when the speaker's tone and voice get louder and more rapid.
Therefore, we extract zero crossing rate (ZCR) and root mean square (RMS) from the original audio tracks to imitate this correlation. Zero Crossing Rate (ZCR) refers to the number of times the speech signal passes through the zero point in each frame. A larger zero-crossing rate is associated with a higher frequency. Root Mean Square (RMS) represents the energy of the speech signal. We observe that in some cases, there exists a weak correlation between ZCR, RMS, and the head pose sequences of listeners, as shown in Fig.~\ref{fig:weak_correlation}.
This correlation improves the prediction performance of the head pose generation module for listeners. 

\textbf{Head Pose Generation.}
With ZCR, RMS, and the corresponding pose sequences from speakers, we intend to obtain predicted head pose sequences consistent with the audio and the head poses of the speaker and listener. TCN~\cite{bai2018empirical} with hierarchical dilated convolutions is suitable for modeling the pose sequences. The structure of hierarchical dilated convolutions enables an exponentially large receptive field, and is powerful to extract multi-scale features of the input sequences. We apply TCN~\cite{bai2018empirical} to extract the rhythm and trend of the audio features and the head poses of the speaker and listener, and then predict the new pose sequences with TCN. The generated head poses are denoted as $\mathbf{f}_{p,s}$ and $\mathbf{f}_{p,l}$ for the speaker and listener, respectively.
\textbf{Expression/Audio Feature of the Speaker.}
We apply the 3DMM~\cite{ploumpis2019combining} face expression parameters as expression features. The 3D face landmark coordinates $S$ can be represented as the combination of 3DMM~\cite{ploumpis2019combining} expression and geometry parameters:
\begin{equation}
    S = \overline{S} + B_{geo} F_{geo} + B_{exp} F_{exp},
    \label{equation:equ0}
\end{equation}
where $\overline{S} \in \mathbb{R}^{3N}$ is the averaged facial mesh, $B_{geo}$ and $B_{exp}$ are the PCA basis of geometry and expression. $F_{geo}$ and $F_{exp}$ are the coefficients of geometry and expression basis. To make the expression feature to be consistent with input audio, we extract the audio features of the speakers like Wav2Lip~\cite{r13}, and then disentangle the expression feature~\cite{yao2022dfa} to be consistent with the input audio.
We take the disentangled feature as the implicit expression/audio feature for the speaker, denoted as $\mathbf{f}_s$. We find that this feature helps learn the eye blinks, and thus helps produce rendering results with more natural visual effects. 

\textbf{Expression Features of the Listener.}
In order to generate natural and realistic reactions for the listeners, we extract the expression features as exemplified by Eq.~\ref{equation:equ0}. More precisely, we exclusively extract features from the frames in which the speaker is actively engaged in speech, thereby enabling the features to more accurately capture the listeners' responses to the corresponding audio input. The expression feature $\mathbf{f}_l$ is utilized as the condition for the neural radiance field as demonstrated in Fig.~\ref{fig:framework}.


\subsection{Dialogue Avatar Portrait Synthesis}
\label{subsec:Avatar Portrait Synthesis}
\textbf{Neural Radiance Field (NeRF) Revisit.} 
Mildenhall \etal~\cite{r12} proposed NeRF, which utilizes a neural network to represent a continuous mapping from a 3D location and a 2D viewing direction to an RGB color value and a volume density value. In particular, NeRF first maps each location $\mathbf{x} \in \mathbb{R}^3$ and viewing direction $\mathbf{d} \in \mathbb{R}^2$ to a high-dimensional embedding space using a sinusoidal positional encoding:
\begin{equation}
  \gamma(p) = (\sin{2^0 \pi p}, \cos{2^0 \pi p}, \sin{2^1 \pi p}, \cos{2^1 \pi p}, \dots).
  \label{eq:1}
\end{equation}
Then a coordinate-based multi-layer perceptron (MLP) $\mathcal{F}_{\theta}$ with parameters $\theta$ is used to map the embedding results to color values $\mathbf{c} \in \mathbb{R}^3$ and volume densities $\sigma \in \mathbb{R}^+$. The overall mapping process is formulated as follows
\begin{equation}
  \mathcal{F}_{\theta} : (\gamma(\mathbf{x}), \gamma(\mathbf{d})) \mapsto (\mathbf{c}, \sigma).
  \label{eq:2}
\end{equation}

With the color $\mathbf{c}$ and density $\sigma$ predicted by $\mathcal{F}_{\theta}$, the sampled density and RGB values are accumulated along the rays cast through each pixel to calculate the output colors. The expected color $\mathit{C}(\mathbf{r})$ of camera ray $\mathbf{r}(t) = \mathbf{o} + t \mathbf{d}$ with camera origin $\mathbf{o}$, camera viewing direction $\mathbf{d}$, near bound $t_n$ and far bound $t_f$ is:
\begin{equation}
  \mathit{C}(\mathbf{r}) = \int_{t_n}^{t_f} T(t) \sigma(\mathbf{r}(t)) \mathbf{c}(\mathbf{r}(t), \mathbf{d}) \mathrm{d}t,
  \label{eq:3}
\end{equation}
where $T(t)$ is the accumulated transmittance along the ray from $t_n$ to $t$:
\begin{equation}
  T(t) = \exp\left(-\int_{t_n}^t \sigma (\mathbf{r}(s)) \mathrm{d}s\right).
  \label{eq:4}
\end{equation}

\textbf{Multi-modal and Individual Conditions.}
The original NeRF model can only represent a static scene, however, talking head generation not only requires a dynamic model, but also needs to control the field with the input audio.
To address this issue, based on standard neural radiance field scene representation, we present a dynamic NeRF conditioned on signals of different modalities, as depicted in Fig.~\ref{fig:framework}.

Specifically, the head pose features ($\mathbf{f}_{p,s}$ and $\mathbf{f}_{p,l}$) are used to determine the position of the camera and the ray casting direction, thus determining the 3D locations $\mathbf{x}$ and the corresponding 2D viewing directions $\mathbf{d}$ for the MLP query. 

For the expressions of individuals, such as mouth movement and eye blinks, we employ different conditional signals for the speaking and listening avatars. Specifically, for the speaking avatar, the conditional signal is the audio features $\mathbf{f}_s$ extracted from the speaker's original audio track. The extracted audio features are processed with a contrastive learning module~\cite{yao2022dfa} to ensure their correspondence with the facial expressions of the speaker.
To construct the face of a listening avatar, we extract the expression feature of the listener as the conditional signal. The extracted expression feature $\mathbf{f}_l$ represents the facial expression of the listener that corresponds to the audio spoken by the speaker.

To model both the speaker and listener in a single model, and meanwhile achieve stronger representation power, inspired by Schwarz \etal~\cite{r19}, we use a pair of latent codes to control the synthesis of an individual, denoted as $\mathbf{z}_{s}$ and $\mathbf{z}_{a}$ respectively. $\mathbf{z}_{s}$ is expected to control the shape of the head, and thus added to $\gamma(\mathbf{x})$ to make a conditional effect on both volume density $\sigma$ and RGB color $\mathbf{c}$. $\mathbf{z}_{a}$ is added to $\gamma(\mathbf{d})$ and only conditions the value of $\mathbf{c}$, which affects the appearance. We have a pair of latent codes for each individual, \ie, $(\mathbf{z}_{a,s}, \mathbf{z}_{s,s})$ for the speaker and $(\mathbf{z}_{a,l}, \mathbf{z}_{s,l})$ for the listener. 
With the help of these latent codes, we are able to build various avatars in a single network, which reduces the computation overhead during training and inference.

In summary, the conditional signal for the speaker is the audio feature and the pose while the conditional signal for the listener is the expression feature and the pose, and a pair of latent codes are learned for each individual. The overall mapping function can be represented as:
\begin{equation}
  \mathcal{F}_{\theta} : (\gamma(\mathbf{x}), \gamma(\mathbf{d}),  \mathbf{f}_s, \mathbf{f}_{p,s}, \mathbf{z}_{s,s}, \mathbf{z}_{a,s}) \mapsto (\mathbf{c}_s, {\sigma}_s), 
\end{equation}
\begin{equation}
       \mathcal{F}_{\theta} : (\gamma(\mathbf{x}), \gamma(\mathbf{d}),  \mathbf{f}_l, \mathbf{f}_{p,l}, \mathbf{z}_{s,l}, \mathbf{z}_{a,l}) \mapsto (\mathbf{c}_l, {\sigma}_l).
\end{equation}

\textbf{Deformation Field for Torso.}
In DialogueNeRF, we construct the head and torso separately in distinct neural radiance fields. When executing volume rendering, sampled points with $\mathbf{c}$ and $\sigma$ values from each ray in head fields are jointed with the sampled points from another ray in torso fields. The two rays are cast through the same pixel in the resulting image. In this way, we can render the head and torso into the same image with one neural volume rendering process. 
Moreover, by concatenating RGB value of the background image to the last point of each ray, the density values in the human part are learned to be high, while the ones in the background part are predicted to be low. Therefore, DialogueNeRF can decouple the foreground and background of the rendered images.

For the torso of both the avatars, the condition signal is the positional encoding of the concatenation of the three-dimensional Euler angle and three-dimensional translation, defined as $\mathbf{et} \in \mathbb{R}^{6}$, where $\mathbf{et}$ is transformed from the head pose in the current frame.
In some situations, the torso in the video moves frequently and randomly, which makes it more challenging to predict its location. In order to construct a torso consistent with the head movement, we adopt a deformation field~\cite{park2021nerfies} to better learn the torso transformation. Specifically, the deformation field receives $\gamma(\mathbf{x})$ as input, and yields a warped location $\mathbf{x'}$. $\mathbf{x'}$ has the same dimension as $\gamma(\mathbf{x})$. It is then concatenated with $\gamma(\mathbf{et})$ and sent to the dynamic neural radiance field for torso synthesis. Finally, we use the L2 reconstruction loss between the rendering outputs and the ground truth videos to supervise the training of our framework.

\begin{table*}[t]
    \small
  \centering
    \caption{Results of comparison. For \cite{r43} and \cite{r8} we use their released demo videos to conduct evaluation. For \cite{r42}, since it does not provide the pre-trained model or demo videos, we train its model with its released code. To ensure the fairness of the experiment, we train our model using the videos employed by Lu \etal~\cite{r43} and Guo \etal~\cite{r8}.}
  \begin{tabular}{c|c|c|c|c|c} 
    \hline
    \textbf{Methods} & \textbf{CPBD} $\uparrow$ & $\textbf{Sync}_{\textbf{conf}}\uparrow$ & \textbf{Resolution} & \textbf{Pose Range} & \textbf{Qualitative Performance} \\
    \hline
    Zhou \etal~\cite{r42} & 0.073 & 0.509 & 128 $\times$ 128  &  small range  &   realistic but not so natural  \\
    \hline
    Lu \etal~\cite{r43} & \textbf{0.338} & 3.125 & 512 $\times$ 512  &  small range, nearly fixed  & real-time and realistic but unnatural \\
    \hline
    Guo \etal~\cite{r8} & 0.172 & \textbf{4.371} &  512 $\times$ 512  & wide range  & realistic but sometimes separated head and torso\\
    \hline
    Ours & 0.328 & 4.032 &  512 $\times$ 512  & wide range  &  realistic and natural \\
    \hline
  \end{tabular}
  \label{tab:compare}
\end{table*}

\begin{table*}[t]
\small
\centering
\caption{Quantitative evaluations on the collected dataset.  }
\label{Tab:Quantitative Evaluations}
\begin{tabular}{c|cccc|cccc}
\hline
\multirow{3}{*}{\textbf{Metrics}} & \multicolumn{4}{c|}{\textbf{YouTube}}                                                                         & \multicolumn{4}{c}{\textbf{CGTN}}                                                                            \\ \cline{2-9} 
                         & \multicolumn{2}{c|}{\textbf{Speaker}}                                & \multicolumn{2}{c|}{\textbf{Listener}}          & \multicolumn{2}{c|}{\textbf{Speaker}}                                & \multicolumn{2}{c}{\textbf{Listener}}          \\ \cline{2-9} 
                         & \multicolumn{1}{c|}{Generated} & \multicolumn{1}{c|}{~~GT~~~}    & \multicolumn{1}{c|}{Generated} & ~~GT~~    & \multicolumn{1}{c|}{Generated} & \multicolumn{1}{c|}{~~GT~~~}    & \multicolumn{1}{c|}{Generated} & ~~GT~~    \\ \hline
CPBD$\uparrow$                     & \multicolumn{1}{c|}{0.453}     & \multicolumn{1}{c|}{~0.493~~} & \multicolumn{1}{c|}{0.308}     & ~0.416~ & \multicolumn{1}{c|}{0.199}     & \multicolumn{1}{c|}{~0.277~~} & \multicolumn{1}{c|}{0.156}     & ~0.284~ \\ \hline
PSNR$\uparrow$                     & \multicolumn{1}{c|}{24.183}    & \multicolumn{1}{c|}{~-~~}     & \multicolumn{1}{c|}{33.464}    & ~-~     & \multicolumn{1}{c|}{36.154}    & \multicolumn{1}{c|}{~-~~}     & \multicolumn{1}{c|}{35.205}    & ~-~     \\ \hline
SSIM$\uparrow$                     & \multicolumn{1}{c|}{0.911}     & \multicolumn{1}{c|}{~-~~}     & \multicolumn{1}{c|}{0.954}     & ~-~     & \multicolumn{1}{c|}{0.968}     & \multicolumn{1}{c|}{~-~~}     & \multicolumn{1}{c|}{0.952}     & ~-~     \\ \hline
Sync$\uparrow$                     & \multicolumn{1}{c|}{7.160}     &
\multicolumn{1}{c|}{~8.296~~} & \multicolumn{1}{c|}{-}         & ~-~     & \multicolumn{1}{c|}{7.895}     & \multicolumn{1}{c|}{~8.828~~} & \multicolumn{1}{c|}{-}         & ~-~     \\ \hline
\end{tabular}
\end{table*}

\section{Experiments}
In this section, we first give our detailed experiment settings. Second, we evaluate our rendering result quality by quantitative and qualitative evaluations. Third, we present the results of our head pose generation method by visualizing the trajectory of the predicted pose sequences with rendered image sequences. Finally, we conduct an ablation study on the pose generation module, the deformation field, and the audio conditions.

\subsection{Evaluation Protocol}
\label{subsec:evaluation}
There are a series of methods for speaking frame generation, together with a number of generation quality metrics. However, to the best of our knowledge, the visual generation of the listener in a communication process has not received enough attention. Therefore there exist few evaluation techniques for the listener. 

\begin{figure}[t]
  \centering
   \includegraphics[width=0.95\linewidth]{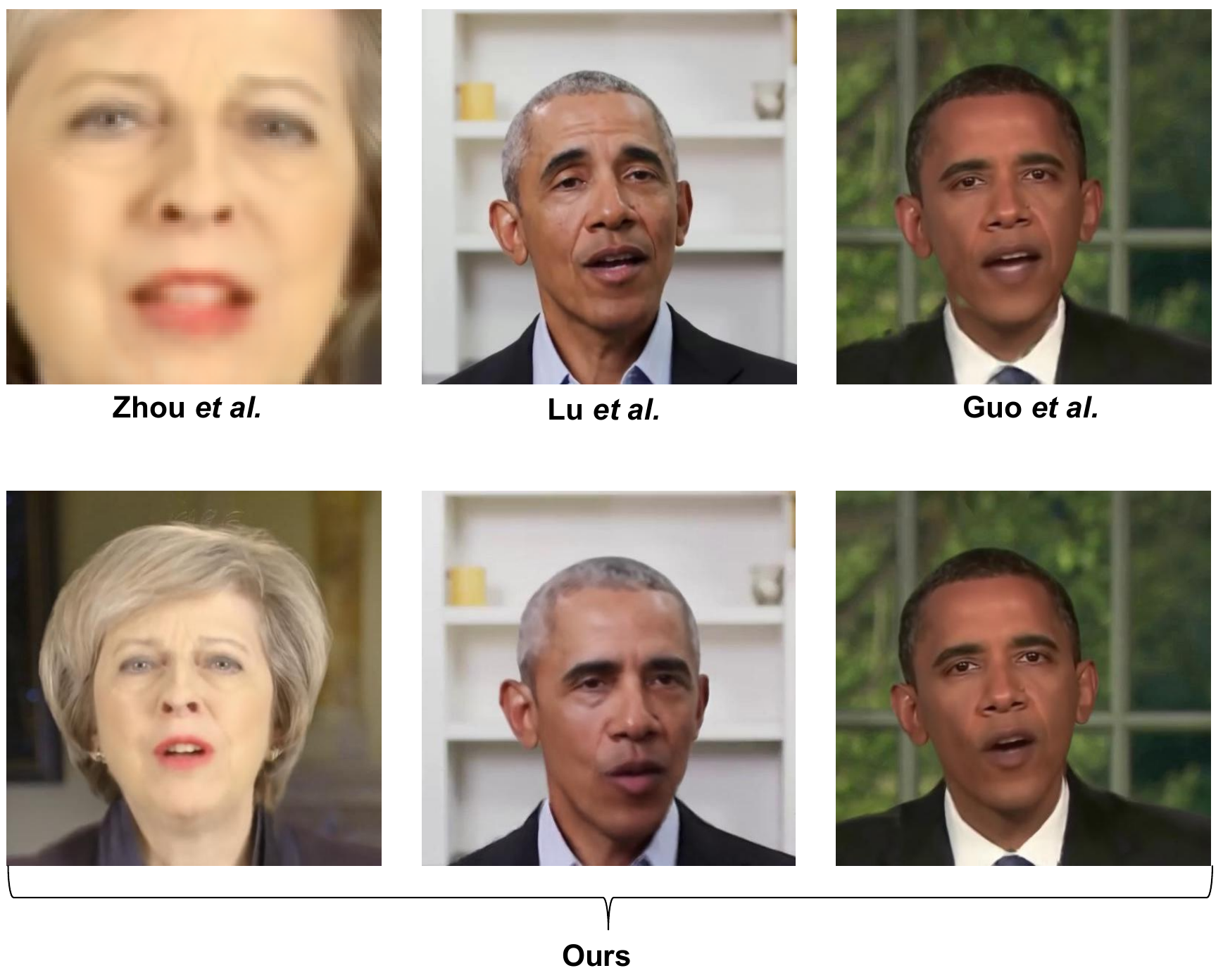}
\vspace{-5pt}
   \caption{Comparison with three methods~\cite{r42, r43, r8}. The images in the second row are from our model. From left to right: we train only the speaker to make a comparison with Zhou \etal~\cite{r42}, Lu \etal~\cite{r43}, and Guo \etal~\cite{r8} respectively. In the first column, Zhou \etal~\cite{r42} renders images with only 128 $\times$ 128 resolution, but our model can produce a 512 $\times$ 512 resolution image. In the second column, Lu \etal~\cite{r43} can only construct a speaking Obama with nearly fixed head poses, however, our model can learn a wider pose range. In the third column, AD-NeRF~\cite{r8} suffers from the separation of head and torso in the reference stage, while our method synthesizes a complete and natural avatar.}
   \label{fig:comparison}
\end{figure}

\textbf{Evaluation Metrics.}
We conduct quantitative evaluations to validate the image rendering quality of DialogueNeRF.
We utilize the cumulative probability blur detection (CPBD)~\cite{r44} to evaluate the sharpness of the resulting images.
Besides, We employ Peak Signal-to-Noise Ratio (PSNR)~\cite{hore2010image} and Structural Similarity (SSIM)~\cite{DBLP:journals/tip/WangBSS04} to compare the image quality with the ground-truth. 
Finally, we utilize the confidence score (Sync) proposed in SyncNet~\cite{r45} to evaluate the accuracy of lip synchronization. 
Since the generated pose sequences cannot be the same as the ground truth poses, when testing PSNR and SSIM we use the poses extracted from the testing videos.

\textbf{Implementation Detail.}
We implement our framework in PyTorch~\cite{r20}. DialogueNeRF is trained with Adam~\cite{r36} optimizer with an initial learning rate of 0.0002 for 600k iterations. In each iteration, we randomly sample a batch of 2048 rays through image pixels. The pose generation module is trained with Adam~\cite{r36} optimizer with an initial learning rate of 0.00005 for 200 epochs. All the experiments are conducted on one NVIDIA RTX3090 GPU. 

\begin{figure}[t]
  \centering
   \includegraphics[width=0.95\linewidth]{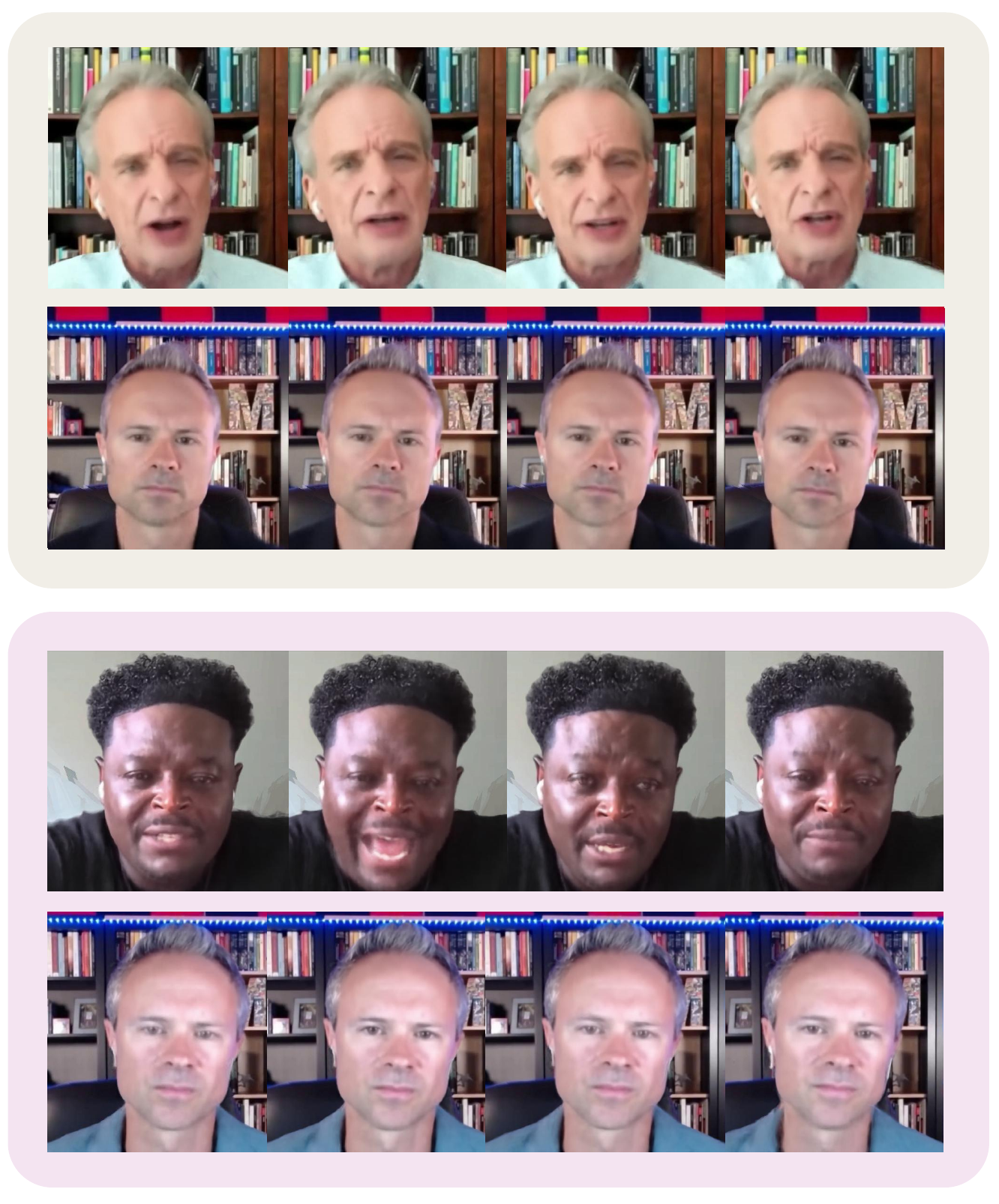}
\vspace{-5pt}
   \caption{Image sequences of different speakers and listeners. While the speaker's poses and expressions vary, most of the time listeners preserve the peace with little pose/expression variations.}
   \label{fig:speaker and listener sequence}
\end{figure}

\subsection{Quantitative and Qualitative Evaluations}
\textbf{Quantitative Evaluations.} 
In the taking head generation scenario, we employ multiple metrics to compare our method with others~\cite{r42, r43, r8}, with results shown in Table~\ref{tab:compare}. Each of these recent methods represents the state-of-the-art among the methods based on similar techniques. Therefore, our comparisons can show the superiority of our model from different aspects. To evaluate the sharpness of the resulting images, we adapt the cumulative probability blur detection (CPBD) metric~\cite{r44}. For the sake of lip synchronization evaluation with input audio, we employ SyncNet~\cite{r45} to calculate the confidence scores to determine the lip-sync error. All the videos are driven by self-audio outside the training dataset. For ATVG~\cite{r42}, we give a simple implementation based on their code. For LSP~\cite{r43} and AD-NeRF~\cite{r8}, we use their released demos. Due to the requirement of SyncNet pre-trained model, we have to resize all the input frames to the size of 224 $\times$ 224 when using SyncNet for evaluation. Furthermore, we list some key characteristics of each method, in order to give a comprehensive comparison. As can be observed, although our framework models both the speaker and listener, DialogueNeRF achieves comparable performance on CPDB and Sync score with the state-of-the-art methods which only model a single speaker. 

In the dialogue video generation scenario, we conduct experiments in videos from our collected YouTube dataset and CGTN dataset, and the results are shown in Table~\ref{Tab:Quantitative Evaluations}. 
Since the image resolutions of the YouTube dataset is larger than that in CGTN dataset, our model achieves better CPBD in the former and better PSNR and SSIM in the latter. We expect these results will set a good baseline for future works devoted to this task.

\begin{table}[t]
\small
\centering
\caption{User study on the performance of the generated conversation videos. }
\label{Tab:user_study}
\begin{tabular}{c|cc|cc}
\hline
\multirow{2}{*}{\textbf{Metrics}}     & \multicolumn{2}{c|}{\textbf{Speaker}}          & \multicolumn{2}{c}{\textbf{Listener}}         \\ 
\cline{2-5} 
                             & \multicolumn{1}{c|}{\textbf{Generated}} &  \textbf{GT} & \multicolumn{1}{c|}{\textbf{Generated}} & \textbf{GT}   \\ \hline
authenticity            & \multicolumn{1}{c|}{2.5}       & 4.93 & \multicolumn{1}{c|}{4}         & 4.78 \\ \hline
 head movement & \multicolumn{1}{c|}{2.79}      & 4.86 & \multicolumn{1}{c|}{3.93}      & 4.71 \\ \hline
eye blink     & \multicolumn{1}{c|}{3}         & 4.92 & \multicolumn{1}{c|}{3.92}      & 4.42 \\ \hline
audio-lip           & \multicolumn{1}{c|}{2}         & 5    & \multicolumn{1}{c|}{-}         & -    \\ \hline
\end{tabular}
\end{table}

\begin{figure}[t]
  \centering
  \includegraphics[width=\linewidth]{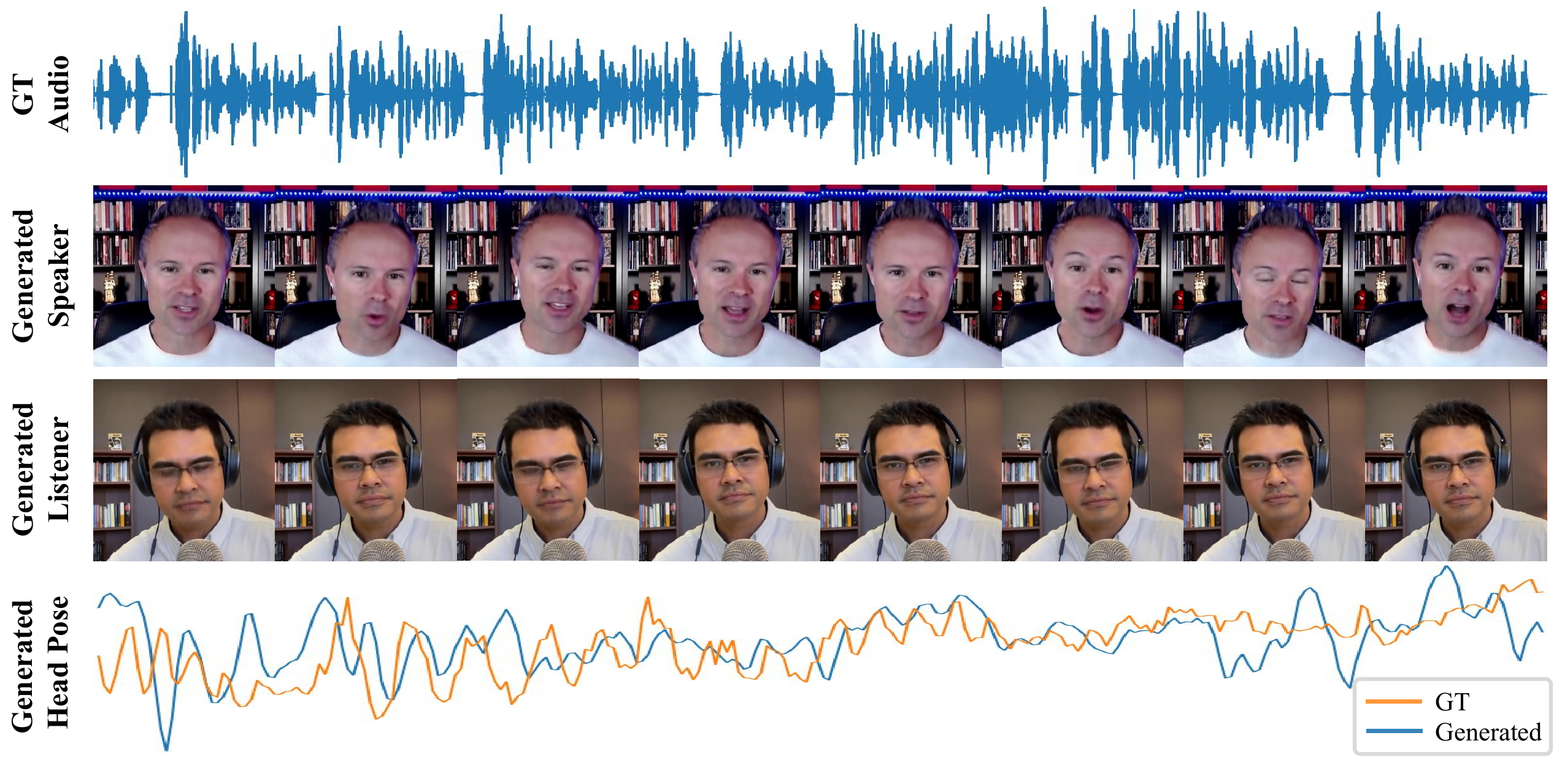}
  \includegraphics[width=\linewidth]{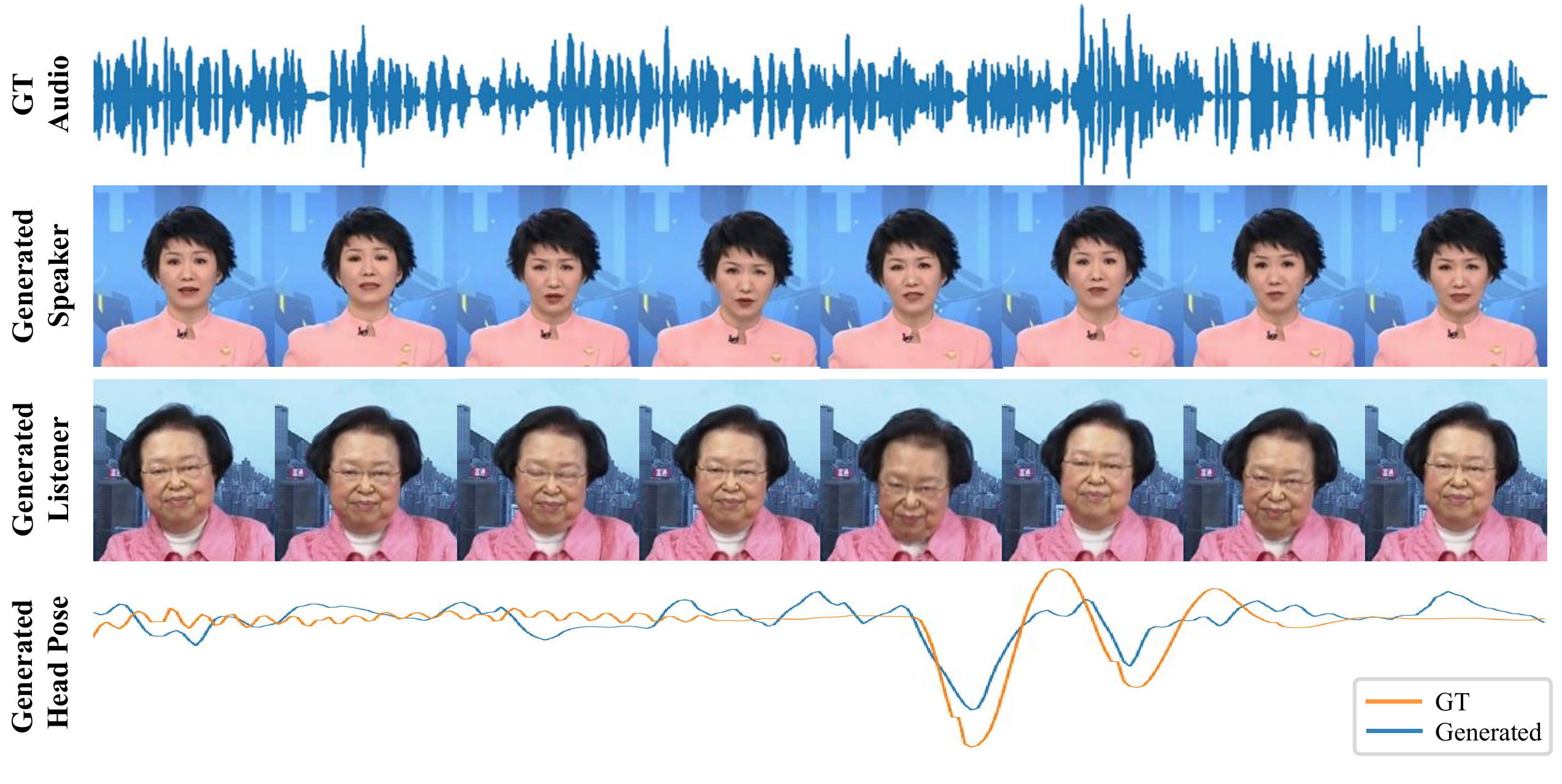}
  \caption{Visualization of the trajectories of listeners' predicted pose sequences with rendered image sequences. 
  The speakers, whose mouth types and head poses are consistent with the input audio, are talking naturally.
  The listeners receive the visual and acoustic information from the speakers and respond with nodding and eye blinking. The generated pose curves (blue lines) show a high degree of consistency with the ground truth (orange lines). }
  \label{fig:pose}
\end{figure}

\begin{figure*}[t]
  \centering
  \includegraphics[width=0.9\linewidth]{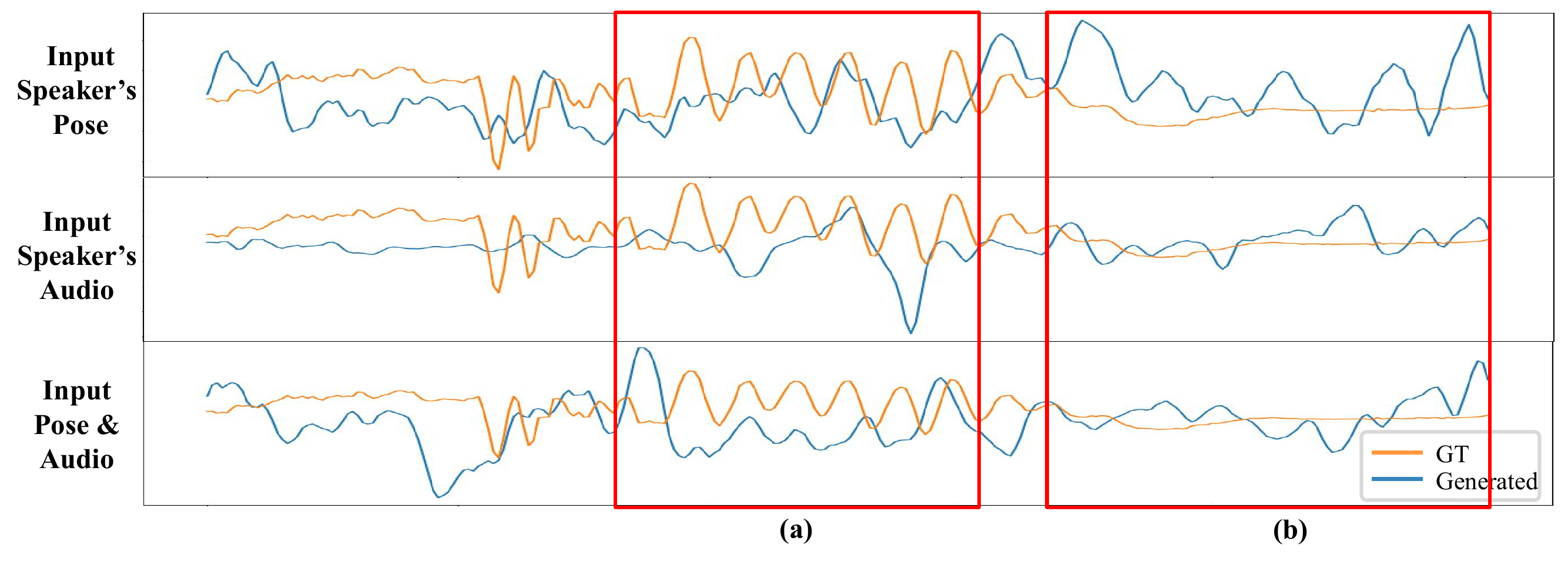}
  
  \caption{Ablation study on the pose generation module. Blue lines refer to listener's generated pose sequences, and orange lines refer to the ground truth.}
  \label{fig:pose1}
\end{figure*}

\begin{figure*}[t]
    \centering
  \includegraphics[width=0.9\linewidth]{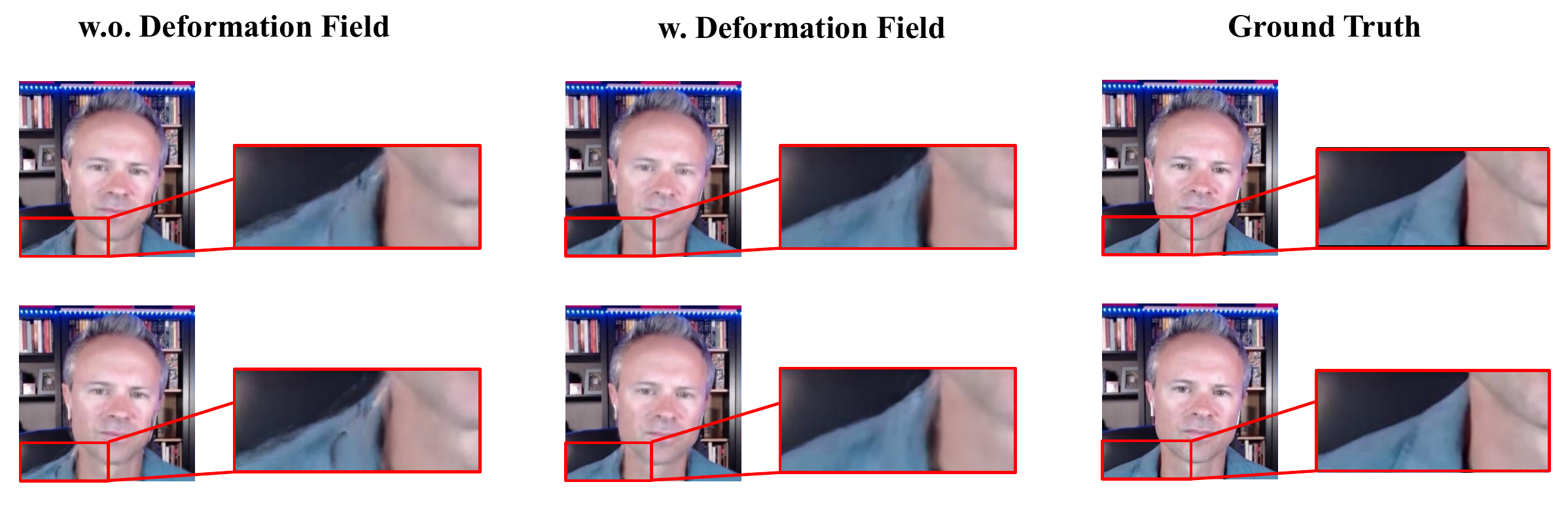}
  \caption{Ablation study on the torso region with or without deformation field. Deformation field alleviates the blurry effects in the boundary of the upper body (\eg, the right shoulder).}
  \label{fig:ablation_torso}
\end{figure*}

\begin{figure}[t]
    \centering
\includegraphics[width=\linewidth]{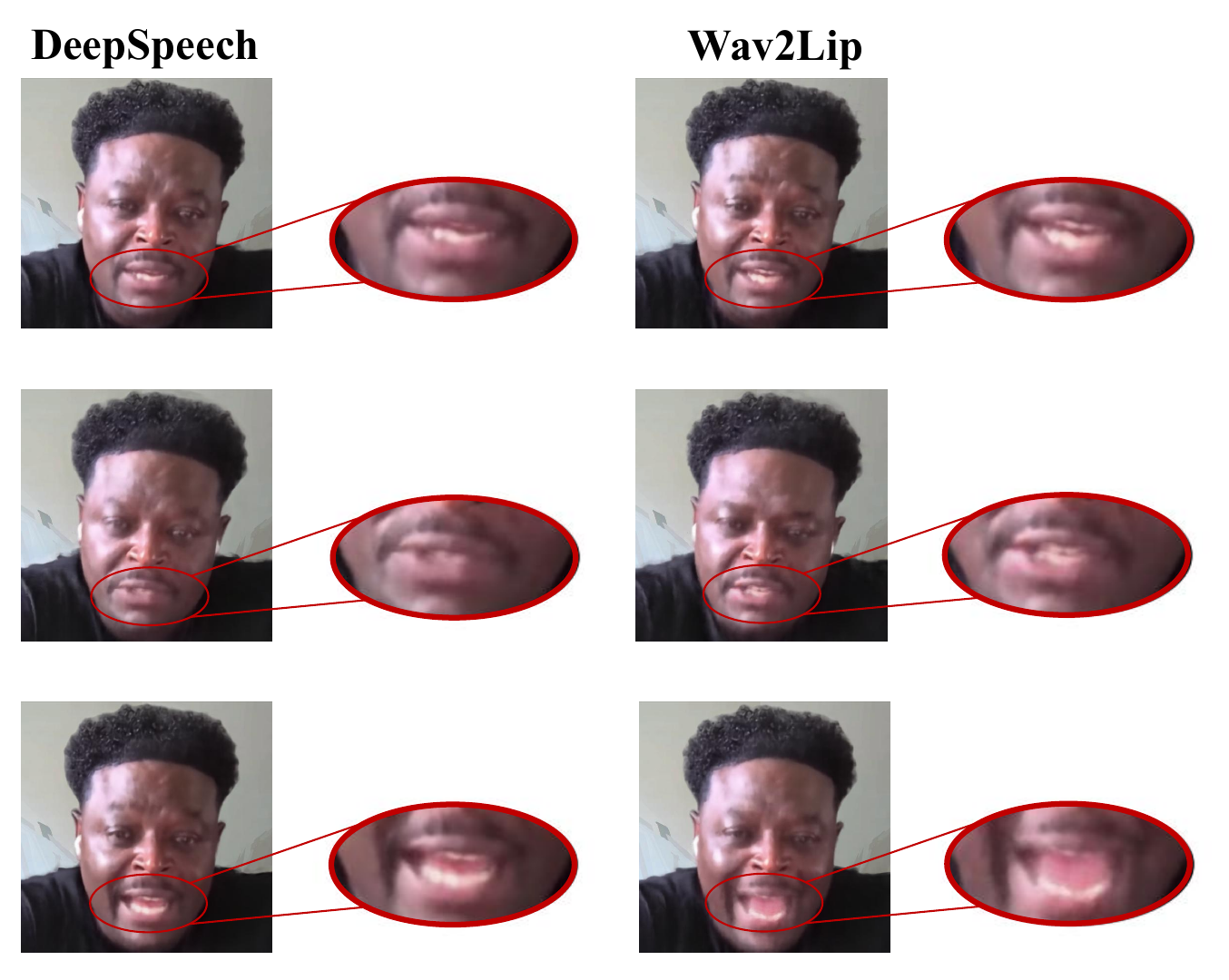}
\vspace{-10pt}
   \caption{Ablation study on conditioning the speaker with {Wav2Lip}~\cite{r13} features and {DeepSpeech}~\cite{r35} features. From the mouth shape, it can be clearly learned that {Wav2Lip} better bridges the domain gap between audio signals and mouth shapes.}
   \label{fig:ablation_mouth}
\end{figure}

\textbf{Qualitative Evaluations.}
In Fig. \ref{fig:comparison} and Fig. \ref{fig:speaker and listener sequence}, we present the synthesis results generated by our method as well as three state-of-the-art methods~\cite{r42, r43, r8}. 
Most of the prior arts focus on audio-driven approaches. ATVG~\cite{r42} achieves a pose-controllable talking face generation, but has the image resolution limits. LSP~\cite{r43} realizes a photo-realistic talking-head animation by an image-to-image translation network and synthesizes high-resolution images in real-time. However, the moving range of their generated head is limited, which looks a little unnatural. AD-NeRF~\cite{r8} models the head and torso separately and achieves a surprisingly realistic speaking style. But in the reference stage, the separation of head and torso sometimes occur. Our model successfully avoids the above problems and synthesizes high-resolution and realistic images with a wide range of head poses. 

We make a user study on the authenticity of the image, naturalness of head movement, naturalness of eye blink, and audio-lip alignment. We adopt the widely used Mean Opinion Scores (MOS)~\cite{xu2011properties} rating protocol. Given the ground truth video and the generated conversation videos containing a talking speaker and a corresponding listener, each participant is asked to rate from 1-5 for the generation results and the ground truth based on the aspects mentioned above. We collect the rating results and compute the average scores. The statistics are shown in Table~\ref{Tab:user_study}. Our generated listener videos acquire slightly lower scores than the ground truth does on the first three aspects, which demonstrates the efficiency of our model to generate listeners. Compared with the listeners, the generated speakers are not good enough. It is probably because the speakers show more drastic changes than listeners in movement. Therefore, our method shows stronger generalization ability in listeners than speakers.




In Fig.~\ref{fig:pose}, we visualize the speaker's audio track, as well as the generated images containing both the speaker and listener with the listener's predicted head pose trajectory. Since the listener's head movements are nodding in most cases, we can represent the listener's head movements with the Euler angle values on the pitch axis. The visualized curves reflect the high degree of similarity between the generated poses and the ground truth.


\subsection{Ablation Study}
\textbf{Pose Generation Module.}
We conduct an ablation study to demonstrate the efficiency of the speaker's head poses and audio features (ZCR and RMS), as shown in Fig.~\ref{fig:pose1}. We respectively remove the speaker's poses and audio features from the input of the head pose generation module, visualize the trajectory of predicted pose sequences and compare the result with both input. Blue lines represent the generated Euler angle values on the pitch axis (corresponding to the nodding of listeners), and orange lines represent the ground truth. Each experiment is trained under the same hyperparameter settings. It is distinguishable that the pose sequences with both input display a more similar rhythm to the ground truth, as shown in Fig. \ref{fig:pose1} (a), and are flatter when the ground truth changes little, as shown in Fig. \ref{fig:pose1} (b).

\textbf{Deformation Field.}
To demonstrate the advantages of employing the deformation field for torso generation, we conduct an ablative experiment by training our DialogueNeRF without the deformation field module.
Fig. \ref{fig:ablation_torso} shows the qualitative comparison between DialogueNeRF with and without the deformation field module.
We see that the position of the torso with violent motions cannot be predicted precisely. Therefore, the boundary of the upper body is blurred (see the zoomed-in regions of the upper body for the rendering image and ground truth). This comparison clearly shows the effectiveness of the deformation field.

\textbf{Audio Feature Condition.} 
As aforementioned in Sec.~\ref{subsec:Modal Signal Generation}, {Wav2Lip}~\cite{r13} features have a strong correspondence with the mouth shape. As a result, it helps to construct a mouth with a wider variety. 
To show the necessity of the Wav2Lip feature, we make an ablation study by comparing the performance of Wav2Lip-based DialogueNeRF and DeepSpeech-based DialogueNeRF. We train these two models with the same hyperparameters. Fig. \ref{fig:ablation_mouth} shows the qualitative comparison of the rendered images from these two models.
We can clearly observe that the mouth shape conditioned by {Wav2Lip} can open wider than that of {DeepSpeech}~\cite{r35}. This indicates that {Wav2Lip} helps the network learn better the correspondence between audio and mouth shape.


\subsection{Potential Applications}
Our model is not only capable of generating new head poses, but also able to imitate the head pose from other videos. It is also suitable for modeling multi-interlocutor scenes. It supports rendering different speakers and listeners separately in the same network, which constructs a multi-talker or multi-listener scene, \eg, a multi-person video conference. Furthermore, our model can be used to assist psychological counseling. Combined with a vocal conversation generation model trained in psychological counseling conversations, our method is able to produce a considerately speaking psychological consultant and an open-eared and responsive listener, which conveniently helps relieve the pain of people having mental disorders at a low cost. Additionally, with the development of large language models, our model can be used to generate virtual avatars for better user experiences by visualizing the procedure of prompting and answering. Since the prompting and answering procedure is naturally a dialogue, our model is able to generate visualization for such process. 
With the aid of our model, the user experience of using such large language models may be improved.


\section{Conclusion and Future Work}
We propose a novel task that considers the human face-to-face conversation process, present a dataset and design a novel method to address this task. With the pose generation module and feature extractors, we can obtain the visual and acoustic signals from speakers to listeners. 
Our framework can generate head pose sequences for speakers and listeners, with only speakers' audio inputted. Furthermore, We can synthesize different realistic conversation fragments within one network using conditional dynamic neural radiance fields. 

In the future, we will further work towards the following directions. 1) Since we do not focus on fast inference, the high-resolution rendering process cannot meet the real-time requirement, but our model can benefit from some concurrent NeRF acceleration works \cite{garbin2021fastnerf, yu2021plenoctrees, lombardi2021mixture} to alleviate this problem. 2) Our method can generate realistic conversation videos, bringing potential ethical issues and negative social impacts. Therefore, face anti-spoofing methods should be developed to identify NeRF-based synthesizing models.

{
\bibliographystyle{IEEEtran}
\bibliography{main}
}

\end{document}